\documentclass[10pt,twocolumn,letterpaper]{article}

\usepackage{cvpr}
\usepackage{times}  
\usepackage{helvet}  
\usepackage{courier}  
\usepackage{url}  
\usepackage{graphicx}  

\usepackage{epsfig}
\usepackage{graphicx}
\usepackage{amsmath}
\usepackage{amssymb}
\usepackage{color}
\usepackage{multirow}
\usepackage{caption}
\usepackage{subcaption}
\usepackage{tabularx} 
\usepackage{algorithm}  
\usepackage{algorithmic} 
\usepackage{flushend}
\usepackage{booktabs}

\usepackage[pagebackref=true,breaklinks=true,letterpaper=true,colorlinks,bookmarks=false]{hyperref}

\cvprfinalcopy 

\def\cvprPaperID{216} 

\ifcvprfinal\fi
\begin{document}

\title{Look Across Elapse: Disentangled Representation Learning and \\ Photorealistic Cross-Age Face Synthesis for Age-Invariant Face Recognition}

\author{\normalsize{Jian~Zhao$^{1,2}$\thanks{Jian Zhao is the corresponding author. Homepage: \url{https://zhaoj9014.github.io/}.}, Yu~Cheng$^{1}$, Yi~Cheng$^{3}$, Yang~Yang$^{1}$, Haochong~Lan$^{4}$, Fang~Zhao$^{5}$, Lin~Xiong$^{3}$, Yan~Xu$^{3}$, Jianshu~Li$^{1}$} \\
		\normalsize{Sugiri Pranata$^{3}$, Shengmei Shen$^{3}$, Junliang~Xing$^{6}$, Hengzhu~Liu$^{2}$, Shuicheng~Yan$^{1,7}$, Jiashi~Feng$^{1}$}\\
	\small{$^{1}$National University of Singapore, $^{2}$National University of Defense Technology} \\
		\small{$^{3}$Panasonic R\&D Center Singapore, $^{4}$Nanyang Technological University} \\ \small{$^{5}$Inception Institute of Artificial Intelligence, $^{6}$Institute of Automation, Chinese Academy of Sciences, $^{7}$Qihoo 360 AI Institute} \\
	{\small  \{zhaojian90, e0321276, yang\_yang, jianshu\}@u.nus.edu, \{yi.cheng, yan.xu, sugiri.pranata, shengmei.shen\}@sg.panasonic.com} \\ {\small lanh0001@e.ntu.edu.sg, \{zhaofang0627, bruinxiongmac\}@gmail.com, jlxing@nlpr.ia.ac.cn, hengzhuliu@nudt.edu.cn, \{eleyans, elefjia\}@nus.edu.sg}}

\maketitle

\begin{abstract}
	Despite the remarkable progress in face recognition related technologies, reliably recognizing faces across ages still remains a big challenge. The appearance of a human face changes substantially over time, resulting in significant intra-class variations. As opposed to current techniques for age-invariant face recognition, which either directly extract age-invariant features for recognition, or first synthesize a face that matches target age before feature extraction, we argue that it is more desirable to perform both tasks jointly so that they can leverage each other. To this end, we propose a deep \textbf{A}ge-\textbf{I}nvariant \textbf{M}odel (AIM) for face recognition in the wild with three distinct novelties. First, AIM presents a novel unified deep architecture jointly performing cross-age face synthesis and recognition in a mutual boosting way. Second, AIM achieves continuous face rejuvenation/aging with remarkable photorealistic and identity-preserving properties, avoiding the requirement of paired data and the true age of testing samples. Third, we develop effective and novel training strategies for end-to-end learning the whole deep architecture, which generates powerful age-invariant face representations explicitly disentangled from the age variation. Moreover, we propose a new large-scale \textbf{C}ross-\textbf{A}ge \textbf{F}ace \textbf{R}ecognition (CAFR) benchmark dataset to facilitate existing efforts and push the frontiers of age-invariant face recognition research. Extensive experiments on both our CAFR and several other cross-age datasets (MORPH, CACD and FG-NET) demonstrate the superiority of the proposed AIM model over the state-of-the-arts. Benchmarking our model on one of the most popular unconstrained face recognition datasets IJB-C additionally verifies the promising generalizability of  AIM in recognizing faces in the wild.
\end{abstract}


\section{Introduction}

\begin{figure}[t]
	\begin{center}
		\includegraphics[width=1\linewidth]{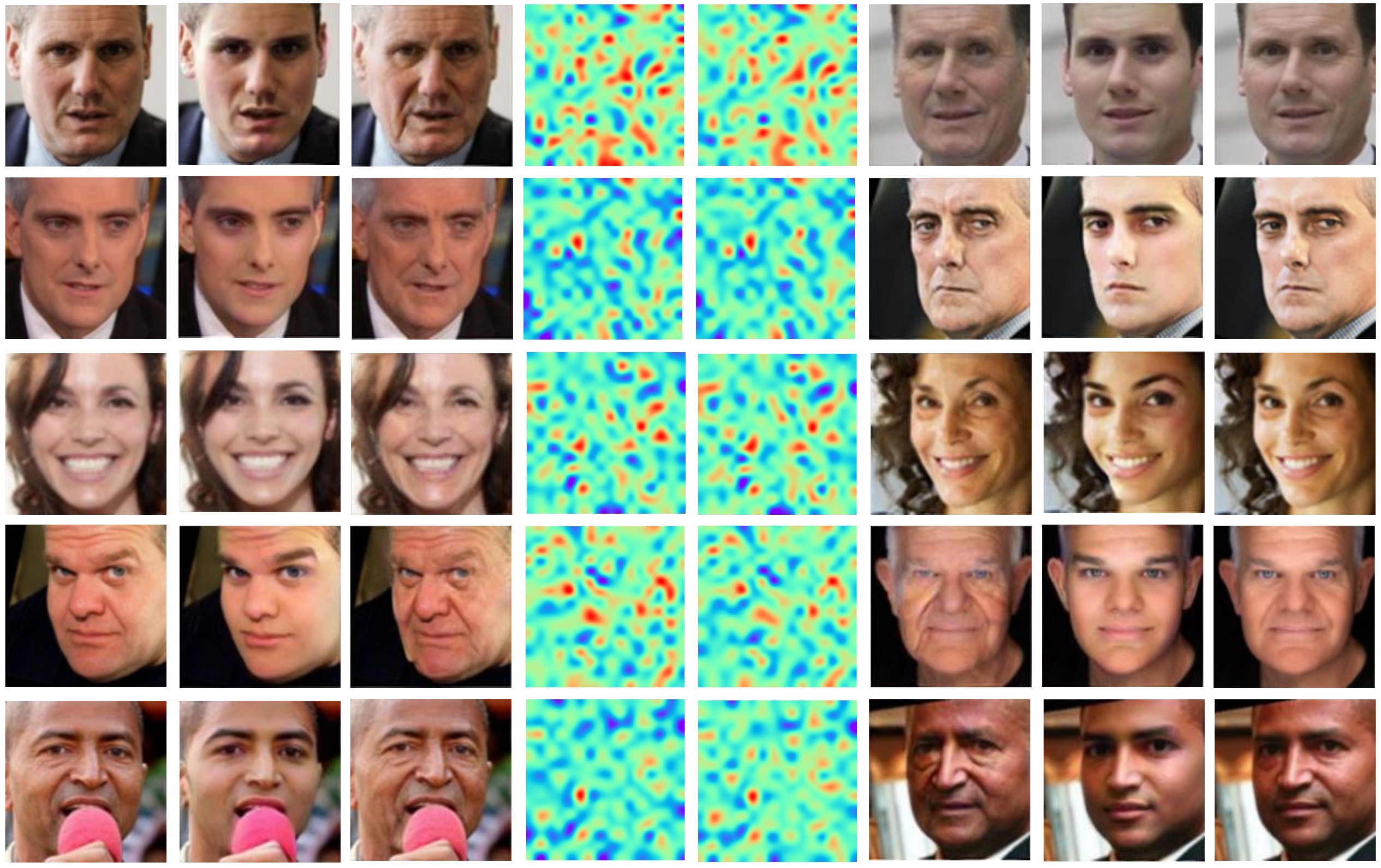}
	\end{center}
	\vspace{-6mm}
	\caption{\small Disentangled Representation Learning and Photorealistic Cross-Age Face Synthesis for Age-Invariant Face Recognition. \emph{Col.} 1 \& 8: Input faces of distinct identities with various challenging factors (\emph{e.g.}, neutral, illumination, expression, pose and occlusion). \emph{Col.} 2 \& 7: Synthesized younger faces by our proposed AIM. \emph{Col.} 3 \& 6: Synthesized older faces by our proposed AIM. \emph{Col.} 4 \& 5: Learned facial representations by our proposed AIM, which are explicitly disentangled from the age variation. AIM can learn age-invariant representations and synthesize photorealistic cross-age faces effectively. Best viewed in color.}
	\label{fig: Figure1}
\end{figure}

Face recognition is one of the most widely studied topics in computer vision and artificial intelligence fields. Recently, some approaches claim to have achieved~\cite{Taigman:Deepface,chen2017robust,li2016robust,zhao2017dual} or even surpassed~\cite{Schroff:Facenet,wang2018cosface,ijcai2018-165} human performance on several benchmarks.

Despite the exciting progress, age variations still form a major bottleneck for many practical applications. For example, in law enforcement scenarios, finding missing children after years, identifying wanted fugitives based on mug shots and verifying passports usually involve recognizing faces across ages and/or synthesizing photorealistic age regressed/progressed\footnote{Face regression (\emph{a.k.a} face rejuvenation) and  face progression (\emph{a.k.a} face aging) refers to  rendering the natural rejuvenation/aging effect for a given face, respectively.} face images. These are extremely challenging due to several reasons:~1) Human face rejuvenation/aging is a complex process whose patterns differ from one individual to another. Both intrinsic factors (like heredity, gender and ethnicity) and extrinsic factors (like environment and living styles) affect the aging process and lead to significant intra-class variations.~2) Facial shapes and textures dramatically change over time, making learning age-invariant patterns difficult.~3) Current learning based cross-age face recognition models are limited by  existing cross-age databases~\cite{fgnet,Rothe-ICCVW-2015,chen2015face,ricanek2006morph,moschoglou2017agedb,zhifei2017cvpr} due to their small size, narrow elapse per subject and unbalanced genders, ethnicities and age span. As such, the performance of most face recognition models degrades by over $13\%$  from general recognition on faces of (almost) the same age to cross-age face recognition~\cite{chen2015face}. In this work, we aim to improve  automatic models for recognizing  unconstrained faces with large age variations.

According to recent studies~\cite{gong2013hidden,wen2016latent},  face images of different individuals usually share common aging characteristics (\emph{e.g.}, wrinkles), and face images of the same individual contain intrinsic features that are relatively stable across ages. Facial representations of a person in the latent space can hence be decomposed into an age-specific component which reflects the aging effect and an identity-specific component which preserves intrinsic identity information. The latter would be invariant to age variations and ideal for   cross-age face recognition when achievable. This finding inspires us to develop a novel and unified deep neural network, termed as \textbf{A}ge \textbf{I}nvariant \textbf{M}odel (AIM). The AIM jointly learns disentangled identity representations that are invariant to age, and photorealistic cross-age face image synthesis that can highlight important latent representations among the disentangled ones end-to-end. Thus they mutually boost each other to achieve   age-invariant face recognition. AIM takes as input face images of arbitrary ages with other potential distracting factors like various illumination, expressions, poses and occlusion. It outputs facial representations invariant to age variations and meanwhile preserves discriminativeness across different identities. As shown in Fig.~\ref{fig: Figure1}, the AIM can learn age-invariant representations and effectively synthesize natural age regressed/progressed faces.

In particular, AIM extends from an auto-encoder based \textbf{G}enerative \textbf{A}dversarial \textbf{N}etwork (GAN) and includes a disentangled \textbf{R}epresentation \textbf{L}earning sub-\textbf{N}et (RLN) and a \textbf{F}ace \textbf{S}ynthesis sub-\textbf{N}et (FSN) for age-invariant face recognition. RLN consists of an encoder and a discriminator that compete with each other to learn discriminative and age-invariant representations. It introduces cross-age domain adversarial training to promote encoded features that are indistinguishable w.r.t. the shift between multi-age domains, and cross-entropy regularization with a label smoothing strategy to constrain cross-age representations with ambiguous separability. The discriminator incorporates dual agents to encourage the representations to be uniformly distributed to smooth the age transformation while preserving identity information. The representations are then concatenated with a continuous age condition code to synthesize age regressed/progressed face images, such that the learned representations are explicitly disentangled from age variations. FSN consists of a decoder and a local-patch based discriminator that compete with each other to synthesize photorealistic cross-age face images. FSN uses an attention mechanism to guarantee robustness to large background complexity and illumination variance. The discriminator incorporates dual agents to add realism to synthesized cross-age faces while forcing the generated faces to exhibit desirable rejuvenation/aging effects.

Moreover, we propose a new large-scale \textbf{C}ross-\textbf{A}ge \textbf{F}ace \textbf{R}ecognition (CAFR) benchmark dataset to facilitate existing efforts and future research on age-invariant face recognition. CAFR contains 1{,}446{,}500 face images from 25{,}000 subjects annotated with age, identity, gender, race and landmark labels. Extensive experiments on both our CAFR and  other standard cross-age datasets (MORPH~\cite{ricanek2006morph}, CACD~\cite{chen2015face} and FG-NET~\cite{fgnet}) demonstrate the superiority of AIM over the state-of-the-arts. Benchmarking AIM on one of the most popular unconstrained face recognition datasets IJB-C~\cite{maze2018iarpa} additionally verifies its promising generalizability in recognizing faces in the wild. Our code and trained models are available at \url{https://github.com/ZhaoJ9014/High_Performance_Face_Recognition/tree/master/src/Look\%20Across\%20Elapse-\%20Disentangled\%20Representation\%20Learning\%20and\%20Photorealistic\%20Cross-Age\%20Face\%20Synthesis\%20for\%20Age-Invariant\%20Face\%20Recognition.TensorFlow}. Our dataset and online demo will be released soon.

Our contributions are summarized as follows.
\vspace{-1mm}
\begin{itemize}
	\setlength\itemsep{0em}
	\item We propose a novel deep architecture unifying cross-age face synthesis and recognition in a mutual boosting way. 
	\item We develop effective  end-to-end training strategies for   the whole deep architecture to generate powerful age-invariant facial representations explicitly disentangled from the age variations.
	\item The proposed model achieves continuous face rejuvenation/aging with remarkable photorealistic and identity-preserving properties, avoiding the requirement of paired data and true age of testing samples.
	
	\item We propose a new large-scale benchmark dataset CAFR to advance the frontiers of age-invariant face recognition research. 
\end{itemize}

\section{Related Work}

\subsection{Age-Invariant Representation Learning}

Conventional approaches often leverage robust local descriptors~\cite{ramanathan2006face,gong2013hidden,sungatullina2013multiview,gong2015maximum,li2016aging} and metric learning~\cite{weinberger2009distance,ling2010face,chen2013blessing} to tackle age variance. For instance, \cite{ramanathan2006face} develop a Bayesian classifier to recognize age difference and perform face verification across age progression.  \cite{gong2013hidden} propose \textbf{H}idden \textbf{F}actor \textbf{A}nalysis (HFA)   for age-invariant face recognition that separates aging variations from identity-specific features.  \cite{weinberger2009distance} improve the performance by distance metric learning.  \cite{ling2010face} propose \textbf{G}radient  \textbf{O}rientation \textbf{P}yramid (GOP) for cross-age face verification. In contrast, deep learning models often handle age variance through using a single age-agnostic or several age-specific models with pooling   and specific loss functions~\cite{wen2016latent,zheng2017age,xu2017age,lin2017cross,wangorthogonal}. For instance,  \cite{cheng2017know} propose an enforced softmax optimization strategy to learn effective and compact deep facial representations with reduced intra-class variance and enlarged inter-class distance.  \cite{wen2016latent} propose a  \textbf{L}atent  \textbf{F}actor guided \textbf{C}onvolutional \textbf{N}eural \textbf{N}etwork (LF-CNN) model  to learn age-invariant deep features.  \cite{zheng2017age} propose an \textbf{A}ge  \textbf{E}stimation guided \textbf{CNN} (AE-CNN) model to separate aging variations from identity-specific features. \cite{wangorthogonal} propose an  \textbf{O}rthogonal \textbf{E}mbedding \textbf{CNN} (OE-CNN) model to decompose deep facial representations into two orthogonal components to represent age- and identity-specific features. 

\subsection{Cross-Age Face Synthesis}

Previous methods   can be roughly divided into
physical modeling based and prototype based. The former approaches model the biological patterns
and physical mechanisms of aging, including muscles~\cite{suo2012concatenational}, wrinkles~\cite{ramanathan2008modeling}, and facial structure~\cite{ramanathan2006modeling}. However, they usually require massive annotated cross-age face data with long elapse per subject  which are hard to collect, and they are computationally expensive. Prototype-based approaches~\cite{burt1995perception,kemelmacher2014illumination} often divide faces into groups by ages and select the average face of each group as the prototype. The differences in prototypes between two age groups are then considered as the aging pattern. However, the aged face generated from the averaged prototype may lose personality information. Most of subsequent approaches~\cite{wang2012combining,yang2016face} are data-driven and do not rely much on the biological prior knowledge, and the aging patterns are learned from training data. Though  improve the results, these methods suffer ghosting artifacts   on the synthesized faces. More recently, deep generative networks are exploited. For instance, \cite{wang2016recurrent} propose a smooth face aging process between neighboring groups by modeling the intermediate transition states with \textbf{R}ecurrent \textbf{N}eural \textbf{N}etwork (RNN).  \cite{zhifei2017cvpr} propose a \textbf{C}onditional \textbf{A}dversarial \textbf{A}uto-\textbf{E}ncoder (CAAE) and  achieve face age regression/progression in a holistic framework. \cite{zhu2018facial} propose a \textbf{C}onditional \textbf{M}ulti-\textbf{A}dversarial \textbf{A}uto-\textbf{E}ncoder with \textbf{O}rdinal
\textbf{R}egression (CMAAE-OR) to predict facial rejuvenation and aging.  \cite{song2018dual} propose a  \textbf{Dual}  \textbf{c}onditional  \textbf{GANs} (Dual cGANs) where the primal
cGAN transforms a face image to other ages based on the age condition, while the dual one learns to invert the task.  

Our model differs from them in following aspects:~1) AIM jointly performs cross-age face synthesis and recognition end-to-end to allow them to mutually boost each other for addressing large age variance in unconstrained face recognition.~2) AIM achieves continuous face rejuvenation/aging with remarkable photorealistic and identity-preserving properties, and do not require paired data and true age of testing samples.~3) AIM generates powerful age-invariant face representations explicitly disentangled from age variations through cross-age domain adversarial training and cross-entropy regularization with a label smoothing strategy. 

\section{Age-Invariant Model}

\begin{figure}[t]
	\begin{center}
		\includegraphics[width=1\linewidth]{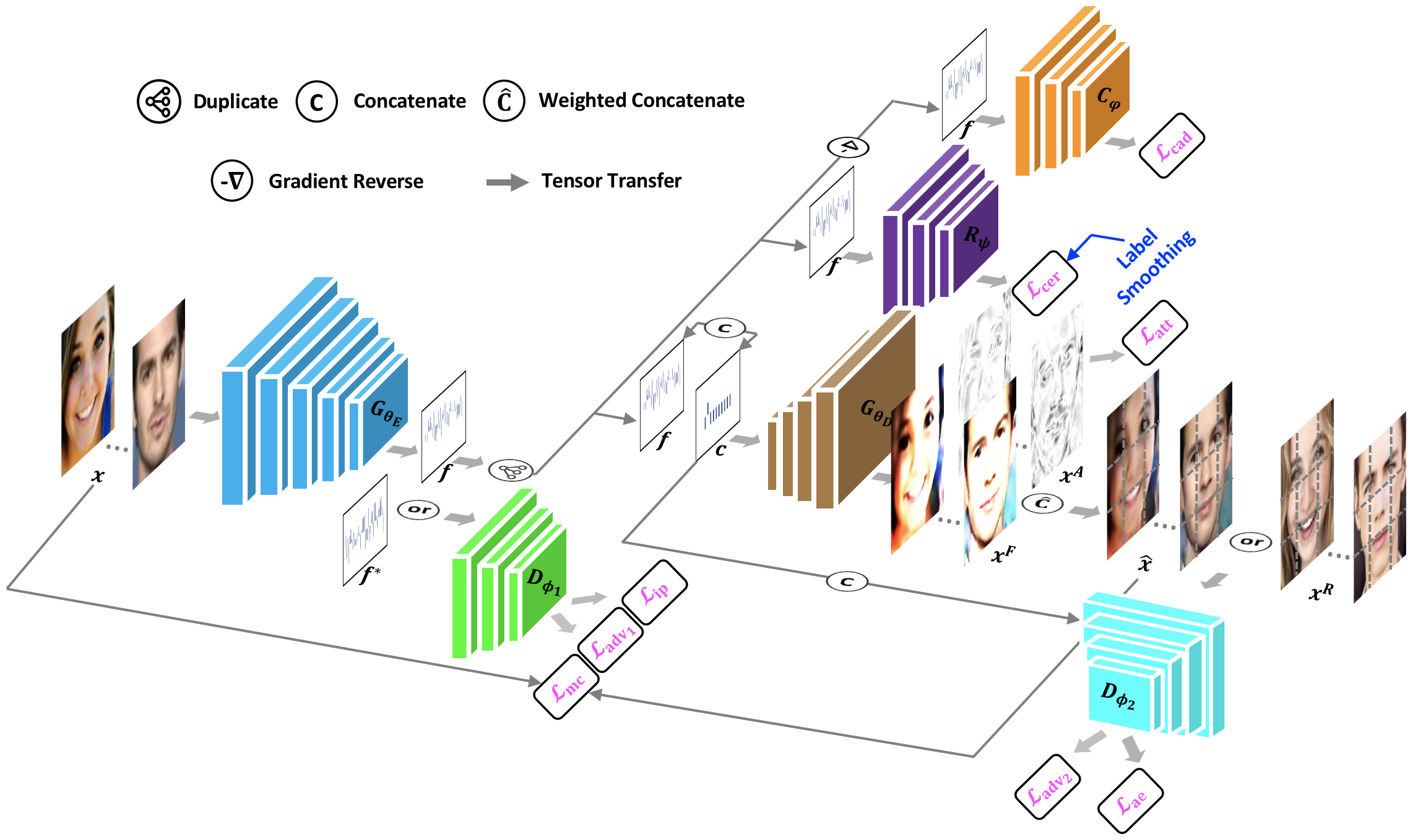}
	\end{center}
	\vspace{-6mm}
	\caption{\small \textbf{A}ge-\textbf{I}nvariant \textbf{M}odel (AIM) for face recognition in the wild. AIM extends from an auto-encoder based GAN and includes a disentangled \textbf{R}epresentation \textbf{L}earning sub-\textbf{N}et (RLN) and a \textbf{F}ace \textbf{S}ynthesis sub-\textbf{N}et (FSN) that jointly learn end-to-end. RLN consists of an encoder ($\mathrm{G}_{\theta_E}$) and a discriminator ($\mathrm{D}_{\phi_1}$) that compete with each other to learn discriminative and robust facial representations ($f$) disentangled from age variance. It is augmented by cross-age domain adversarial training ($\mathcal{L}_\mathrm{cad}$) and cross-entropy regularization with a label smoothing strategy ($\mathcal{L}_\mathrm{cer}$). FSN consists of a decoder ($\mathrm{G}_{\theta_D}$) and a local-patch based discriminator ($\mathrm{D}_{\phi_2}$) that compete with each other to achieve continuous face rejuvenation/aging ($\hat{x}$) with remarkable photorealistic and identity-preserving properties. It introduces an attention mechanism to guarantee robustness to large background complexity and illumination variance. Note AIM does not require paired training data nor true age of testing samples. Best viewed in color.}
	\label{fig: Figure2}
\end{figure}

As shown in Fig.~\ref{fig: Figure2}, the proposed  \textbf{A}ge-\textbf{I}nvariant \textbf{M}odel (AIM) extends from an auto-encoder based GAN, and consists of a disentangled \textbf{R}epresentation \textbf{L}earning sub-\textbf{N}et (RLN) and a \textbf{F}ace \textbf{S}ynthesis sub-\textbf{N}et (FSN) that jointly learn discriminative and robust facial representations disentangled from age variance and perform attention-based face rejuvenation/aging end-to-end. We now detail each component.

\subsection{Disentangled Representation Learning}

Matching face images across ages is demanded in many real-world applications. It is mainly challenged by variations of an individual at different ages (\emph{i.e.} large intra-class variations) or caused by aging (\emph{e.g.} facial shape and texture changes), and inevitable entanglement of unrelated (statistically independent) components in the deep features extracted from a general-purpose face recognition model. Large intra-class variations usually result in erroneous cross-age face recognition and entangled facial representations potentially weaken the model's robustness in recognizing faces with age variations. We propose a GAN-like \textbf{R}epresentation \textbf{L}earning sub-\textbf{N}et (RLN) to learn discriminative and robust identity-specific facial representations disentangled from age variance, as illustrated in Fig.~\ref{fig: Figure2}.

In particular, RLN takes the encoder $\mathrm{G}_{\theta_E}$ (with learnable parameters $\theta_E$) as the generator : $\mathbb{R}^{H \times W \times C} \mapsto \mathbb{R}^{C{'}}$ for facial representation learning, where $H$, $W$, $C$ and $C{'}$ denote the input image height, width, channel number and the dimensionality of the encoded feature $f$, respectively. $f$ preserves the high-level identity-specific information of the input face image through several carefully designed regularizations. We further concatenate $f$ with a continuous age condition code to synthesize age regressed/progressed face images, such that the learned representations are explicitly disentangled from age variations.

Formally, denote the input RGB face image as $x$ and the learned facial representation as $f$. Then
\begin{equation}
	\small
	f := \mathrm{G}_{\theta_E}(x).
\end{equation}

The key requirements for $\mathrm{G}_{\theta_E}$ include three aspects.~1) The learned representation $f$ should be invariant to age variations and also well preserve the identity-specific component.~2) It should
be barely possible for an algorithm to identify the domain of origin of the observation $x$ regardless of the underlying gap between multi-age domains. ~3) $f$ should obey uniform distribution to smooth the age transformation. 

To this end, we propose to learn $\theta_E$ by minimizing the following composite losses:
\begin{equation}
	\begin{aligned}
		\small
		\mathcal{L}_{\mathrm{G}_{\theta_E}}&=-\lambda_1 \mathcal{L}_\mathrm{cad}+\lambda_2 \mathcal{L}_\mathrm{cer}-\lambda_3 \mathcal{L}_\mathrm{{adv}_1}+\lambda_4 \mathcal{L}_\mathrm{{ip}}\\&-\lambda_5 \mathcal{L}_\mathrm{{adv}_2}+\lambda_6 \mathcal{L}_\mathrm{{ae}}+\lambda_7 \mathcal{L}_\mathrm{{mc}}+\lambda_8 \mathcal{L}_\mathrm{{tv}}+\lambda_9 \mathcal{L}_\mathrm{{att}},&
	\end{aligned}
\end{equation}
where $\mathcal{L}_\mathrm{cad}$ is the \textbf{c}ross-\textbf{a}ge \textbf{d}omain adversarial loss for facilitating age-invariant representation learning via domain adaption, $\mathcal{L}_\mathrm{cer}$ is the \textbf{c}ross-\textbf{e}ntropy \textbf{r}egularization loss for constraining cross-age representations with ambiguous separability, $\mathcal{L}_\mathrm{{adv}_1}$ is the \textbf{adv}ersarial loss for imposing the uniform distribution on $f$, $\mathcal{L}_\mathrm{{ip}}$ is the \textbf{i}dentity \textbf{p}reserving loss for preserving identity information, $\mathcal{L}_\mathrm{{adv}_2}$ is the \textbf{adv}ersarial loss for adding realism to
the synthesized images and alleviating artifacts, $\mathcal{L}_\mathrm{{ae}}$ is the \textbf{a}ge \textbf{e}stimation loss for forcing the synthesized faces to exhibit desirable rejuvenation/aging effect, $\mathcal{L}_\mathrm{{mc}}$ is the \textbf{m}anifold \textbf{c}onsistency loss for encouraging input-output space manifold consistency, $\mathcal{L}_{\mathrm{tv}}$ is the \textbf{t}otal \textbf{v}ariation loss for reducing spiky artifacts, $\mathcal{L}_\mathrm{{att}}$ is the \textbf{att}ention loss for facilitating robustness enhancement via an attention mechanism, and $\{\lambda_k\}_{k=_1^9}$ are weighting parameters among different losses.

In order to enhance the age-invariant representation learning capacity, 
we adopt
$\mathcal{L}_\mathrm{cad}$ to promote emergence of features encoded by $\mathrm{G}_{\theta_E}$ that are indistinguishable w.r.t. the shift between
multi-age domains, which is defined as 
\begin{equation}
	\small
	\mathcal{L}_\mathrm{cad}=\frac{1}{N}\sum_i-y_ilog[\mathrm{C}_\varphi(f_i)]-(1-y_i)log[1-\mathrm{C}_\varphi(f_i)],
\end{equation}
where $\varphi$ denotes the learnable parameters for the domain classifier, and $y_i\in\{0,1,\dots\}$ indicates which domain $f_i$ is from. Minimizing $\mathcal{L}_{\mathrm{cad}}$ can reduce the domain  discrepancy and help the generator achieve similar facial representations across different age domains, even if training samples from a domain are limited. Such adapted representations are provided by augmenting the encoder of $\mathrm{G}_{\theta_E}$ with a few standard layers as the domain classifier $\mathrm{C}_{\varphi}$, and a new gradient reversal layer to reverse the gradient during optimizing the encoder (\emph{i.e.}, gradient reverse operator as in Fig.~\ref{fig: Figure2}), as inspired by~\cite{ganin2016domain}.

If using $\mathcal{L}_{\mathrm{cad}}$ alone,  the results tend to be sub-optimal, because searching for a local minimum of $\mathcal{L}_{\mathrm{cad}}$ may go through a path that resides outside the manifold of desired cross-age representations with ambiguous separability. Thus, we combine $\mathcal{L}_{\mathrm{cad}}$ with $\mathcal{L}_{\mathrm{cer}}$ to ensure  the search resides in that manifold and produces age-invariant facial representations, where $\mathcal{L}_{\mathrm{cer}}$ is defined as
\begin{equation}
	\small
	\mathcal{L}_\mathrm{cer}=\frac{1}{N}\sum_i-\bar{y}_ilog[\mathrm{R}_\psi(f_i)]-(1-\bar{y}_i)log[1-\mathrm{R}_\psi(f_i)],
\end{equation}
where $\psi$ denotes the learnable parameters for the regularizer, and $\bar{y}_i\in\{\frac{1}{n},\frac{1}{n},\dots\}$ denotes the smoothed domain indicator. 

$\mathcal{L}_{\mathrm{adv}_1}$ is introduced to impose a prior distribution (\emph{e.g.}, uniform distribution) on $f$ to evenly populate the latent space with no apparent ``holes", such that smooth age transformation can be achieved:
\begin{equation}
	\small
	\mathcal{L}_{\mathrm{adv}_1}=\frac{1}{N}\sum_i-y_ilog[\mathrm{D}_{\phi_1}(f_i)]-(1-y_i)log[1-\mathrm{D}_{\phi_1}(f^*_i)],
\end{equation}
where $\phi_1$ denotes the learnable parameters for the discriminator, $f^*_i\sim U(f)$ denotes a random sample from uniform distribution $U(f)$, and $y_i$ denotes the binary distribution indicator.

To facilitate this process, we leverage a \textbf{M}ulti-\textbf{L}ayer \textbf{P}erceptron (MLP) as the discriminator $\mathrm{D}_{\phi_1}$, which is very simple to avoid typical GAN tricks. We further augment $\mathrm{D}_{\phi_1}$ with an auxiliary agent $\mathcal{L}_\mathrm{ip}$ to preserve identity information:
\begin{equation}
	\small
	\mathcal{L}_\mathrm{ip}=\frac{1}{N}\sum_i-y_ilog[\mathrm{D}_{\phi_1}(f_i)]-(1-y_i)log[1-\mathrm{D}_{\phi_1}(f_i)],
\end{equation}
where $y_i$ denotes the identity ground truth.

\subsection{Attention-based Face Rejuvenation/Aging}

Photorealistic cross-age face images are important for face recognition with large age variance. A natural scheme is to generate reference age regressed/progressed faces from face images of arbitrary ages to match target age before feature extraction or serve as augmented data for learning discriminative models. We then propose a GAN-like \textbf{F}ace \textbf{S}ynthesis sub-\textbf{N}et (FSN) to learn a synthesis function that can achieve both face rejuvenation and aging in a holistic, end-to-end manner, as illustrated in Fig.~\ref{fig: Figure2}.

In particular, FSN leverages the decoder $\mathrm{G}_{\theta_D}$ (with learnable parameters $\theta_D$) as the generator: $\mathbb{R}^{C{'}+C{''}} \mapsto \mathbb{R}^{H \times W \times C}$ for cross-age face synthesis, where $C{''}$ denotes the dimensionality of the age condition code concatenated with $f$. The synthesized results present natural effects of  rejuvenation/aging with robustness to large background complexity and bad lighting conditions through the carefully designed learning schema.

Formally, denote the age condition code as $c$ and the synthesized face image as $\hat{x}$. Then
\begin{equation}
	\small
	\hat{x}:=\mathrm{G}_{\theta_D}(f,c).
\end{equation}

The key requirements for $\mathrm{G}_{\theta_D}$ include two aspects.~1) The synthesized face image $\hat{x}$ should visually resemble a real one and preserve the
desired rejuvenation/aging effect.~2) Attention should be paid to the most salient regions of the image that are responsible for synthesizing the novel aging phase while keeping the rest elements such as glasses, hats, jewelery and background untouched.

To this end, we propose to learn $\theta_D$ by minimizing the following composite losses:
\begin{equation}
	\small
	\mathcal{L}_{\mathrm{G}_{\theta_D}}=-\lambda_{10} \mathcal{L}_{\mathrm{adv}_2}+\lambda_{11} \mathcal{L}_\mathrm{ae}+\lambda_{12} \mathcal{L}_\mathrm{mc}+\lambda_{13} \mathcal{L}_\mathrm{tv}+\lambda_{14} \mathcal{L}_\mathrm{att},
\end{equation}
where $\{\lambda_k\}_{k=_{10}^{14}}$ are weighting parameters among different losses.

$\mathcal{L}_{\mathrm{adv}_2}$ is introduced to push the
synthesized image to reside in the manifold of photorealistic age regressed/progressed face images, prevent blur effect, and produce visually pleasing results:
\begin{equation}
	\scriptsize
	\mathcal{L}_{\mathrm{adv}_2}=\frac{1}{N}\sum_i-y_ilog[\mathrm{D}_{\phi_2}(\hat{x}_i,c_{i,j})]-(1-y_i)log[1-\mathrm{D}_{\phi_2}(x^R_i,c_{i,j})],
\end{equation}
where $\phi_2$ denotes the learnable parameters for the discriminator, $c_{i,j}$ denotes the age condition code to transform $f_i$ into the $j^{th}$ age phase, and $x^R_i$ denotes a real face image with (almost) the same age with $\hat{x}_i$ (not necessarily belong to the same person).

To facilitate this process, we modify a CNN backbone as a local-patch based discriminator $\mathrm{D}_{\phi_2}$ to prevent $\mathrm{G}_{\theta_D}$ from over-emphasizing certain image features to fool the current discriminator network. We further augment $\mathrm{D}_{\phi_2}$ with an auxiliary agent $\mathcal{L}_\mathrm{ae}$ to preserve the desired rejuvenation/aging effect. In this way, $\mathrm{G}_{\theta_D}$ not only learns to render photorealistic samples but also learns to satisfy the target age encoded by $c$:
\begin{equation}
	\small
	\mathcal{L}_\mathrm{ae}=\frac{1}{N}\sum_i\|\hat{c}_{i,j}-c_{i,j}\|^2_2+\|c^R_{i,j}-c_{i,j}\|^2_2,
\end{equation}
where $\hat{c}_{i,j}$ and $c^R_{i,j}$ denote the estimated ages from $\hat{x}_i$ and $x^R_i$, respectively.

$\mathcal{L}_\mathrm{mc}$ is introduced to enforce the manifold consistency between the input-output space, defined as $\|\hat{x}-x\|^2_2 /|x|$, where $|x|$ is the size of $x$. $\mathcal{L}_\mathrm{TV}$ is introduced as a regularization term on the synthesized results to reduce spiky artifacts:
\begin{equation}
	\small
	\mathcal{L}_\mathrm{TV}=\sum_{i,j}^{H,W}\|\hat{x}_{i,j+1}-\hat{x}_{i,j}\|^2_2+\|\hat{x}_{i+1,j}-\hat{x}_{i,j}\|^2_2.
\end{equation}

In order to make the model focus on the most relevant features, we adopt $\mathcal{L}_\mathrm{att}$ to facilitate robustness enhancement via an attention mechanism:
\begin{equation}
	\small
	\mathcal{L}_\mathrm{att}=\sum_{i,j}^{H,W}\|x^A_{i,j+1}-x^A_{i,j}\|^2_2+\|x^A_{i+1,j}-x^A_{i,j}\|^2_2+\|x^A_{i,j}\|^2_2,
\end{equation}
where $x^A$ denotes the attention score map which serves as the guidance, and attends to the most relevant regions during cross-age face synthesis.

The final synthesized results can be obtained by
\begin{equation}
	\small
	\hat{x}=x^A\cdot x^F+(1-x^A)\cdot x,
\end{equation}
where $x^F$ denotes the feature map predicted by the last fractionally-strided convolution block. 

\subsection{Training and Inference}

The goal of AIM is to use sets of real targets to learn
two GAN-like sub-nets that mutually boost each other and jointly accomplish age-invariant face recognition. Each separate loss serves as a deep supervision within the hinged structure benefiting network convergence. The overall objective function for AIM is
\begin{equation}
	\begin{aligned}
		\small
		\mathcal{L}_{\mathrm{AIM}}&=-\lambda_1 \mathcal{L}_\mathrm{cad}+\lambda_2 \mathcal{L}_\mathrm{cer}-\lambda_3 \mathcal{L}_\mathrm{{adv}_1}+\lambda_4 \mathcal{L}_\mathrm{{ip}}\\&-\lambda_5 \mathcal{L}_\mathrm{{adv}_2}+\lambda_6 \mathcal{L}_\mathrm{{ae}}+\lambda_7 \mathcal{L}_\mathrm{{mc}}+\lambda_8 \mathcal{L}_\mathrm{{tv}}+\lambda_9 \mathcal{L}_\mathrm{{att}}.&
	\end{aligned}
\end{equation}

During testing, we simply feed the input face image $x$ and desired age condition code $c$ into AIM to obtain the disentangled age-invariant representation $f$ from $\mathrm{G}_{\theta_E}$ and the synthesized age regressed/progressed face image $\hat{x}$ from $\mathrm{G}_{\theta_D}$. Example results are visualized in Fig.~\ref{fig: Figure1}. 

\section{Cross-Age Face Recognition Benchmark}

\begin{figure}[t]
	\begin{center}
		\includegraphics[width=1\linewidth]{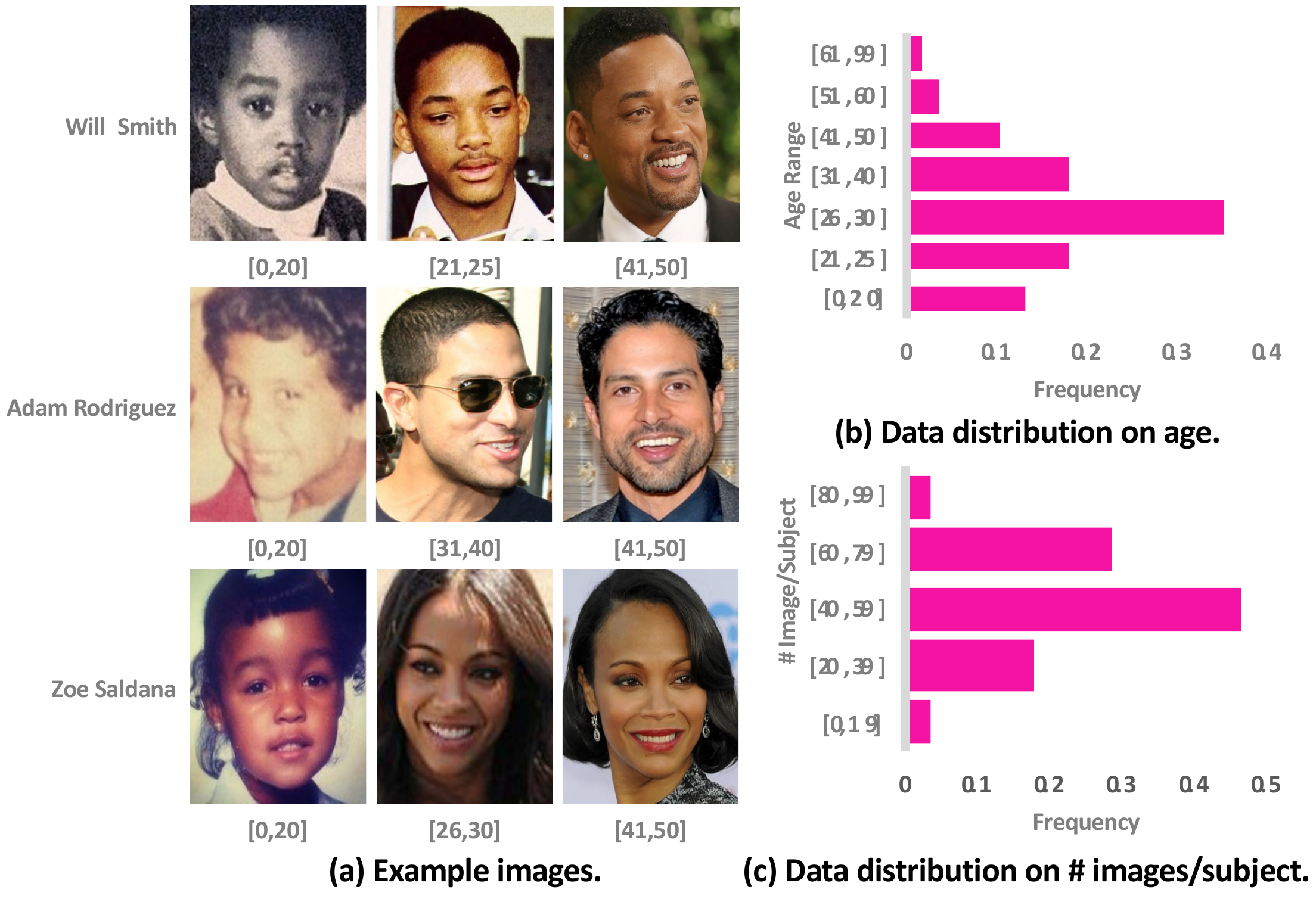}
	\end{center}
	\vspace{-6mm}
	\caption{\small \textbf{C}ross-\textbf{A}ge \textbf{F}ace \textbf{R}ecognition (CAFR) dataset. Best viewed in color.}
	\label{fig: Figure3}
\end{figure}

\begin{table*}[t]
	\newcommand{\tabincell}[2]{\begin{tabular}{@{}#1@{}}#2\end{tabular}}
	\begin{center}
		\small
		\caption{\small Statistics for publicly available cross-age face datasets.}
		\label{tab: Table1}
		\vspace{-2mm}
		\begin{tabular}{cccccc}
			\hline
			{\textbf{Dataset}} & {\textbf{\# Images}} & {\textbf{\# Subjects}} & {\textbf{\# Images/Subject}} & {\textbf{Age Span}} & {\textbf{Average Age}} \\
			\hline
			FG-NET~\cite{fgnet} & 1,002 & 82 & \emph{avg.} 12.22 & 0-69  & 15.84 \\
			MORPH Album1~\cite{ricanek2006morph} & 1,690 & 515 & \emph{avg.} 3.28 & 15-68  & 27.28 \\
			MORPH Album2~\cite{ricanek2006morph} & 78,207 & 20,569 & \emph{avg.} 3.80 & 16-99  & 32.69 \\
			CACD~\cite{chen2015face} & 163,446 & 2,000 & \emph{avg.} 81.72 & 16-62  & 38.03 \\
			IMDB-WIKI~\cite{rothe2015dex} & 523,051 & 20,284 & \emph{avg.} 25.79 & 0-100  & 38.00 \\
			AgeDB~\cite{moschoglou2017agedb} & 16,488 & 568 & \emph{avg.} 29.03 & 1-101  & 50.30 \\
			CAF~\cite{wangorthogonal} & 313,986 & 4,668 & \emph{avg.} 67.26 & 0-80  & 29.00 \\
			\hline
			\textbf{CAFR} & \textbf{1,446,500} & \textbf{25,000} & \textbf{\emph{avg.} 57.86} & \textbf{0-99} & \textbf{28.23} \\
			\hline
		\end{tabular}
	\end{center}
\end{table*}

In this section, we introduce a new large-scale ``\textbf{C}ross-\textbf{A}ge \textbf{F}ace \textbf{R}ecognition (CAFR)" benchmark dataset to push the frontiers of age-invariant face recognition research with several appealing properties.~1) It contains 1,446,500 face images from 25,000 subjects annotated with age, identity, gender, race and landmark labels, which is larger and more comprehensive than previous similar attempts~\cite{fgnet,ricanek2006morph,chen2015face,rothe2015dex,moschoglou2017agedb,wangorthogonal}.~2) The images within CAFR are collected from real-world scenarios, involving humans with various expressions, poses, occlusion and resolution.~3) The background of images in CAFR is more complex and diverse than previous datasets. Some examples and statistics w.r.t. data distribution on the image number per age phase and the image number per subject are illustrated in Fig.~\ref{fig: Figure3} (a), (b) and (c), respectively.

\subsection{Image Collection and Annotation}

We select a sub-set from the celebrity name list of MS-Celeb-1M~\cite{guo2016ms} for data collection based on below considerations.~1) Each individual must have many cross-age face images available on the Internet for retrieval.~2) Both gender balance and racial diversity should be considered. Accordingly, we manually specify some keywords (such as name, face image, event, year, \textit{etc.}) to ensure the accuracy and diversity of returned results. Based on these specifications, corresponding cross-age face images are located by performing Internet searches over Google and Bing image search engines. For each identified image, the corresponding URL is stored in a spreadsheet. Automated scrapping software is used to download the cross-age imagery and stores all relevant information (\emph{e.g.}, identity) in a  database. Moreover, a pool of self-collected children face images with age variations is also constructed to augment and complement Internet scraping results.

After curating the imagery, semi-automatic annotation is conducted with three steps.~1) Data cleaning. We perform face detection with an off-the-shelf algorithm~\cite{li2017towards} to filter the images without any faces and manually wipe off duplicated images and false positive images (\emph{i.e.}, faces that do not belong to that subject).~2) Data annotation. We combine the prior information on identity and apply off-the-shelf age estimator~\cite{rothe2015dex} and landmark localization algorithm~\cite{li2017integrated} to annotate the ground truths on age, identity, gender, race and landmarks.~3) Manual inspection. After annotation, manual inspection is performed on all images and corresponding annotations to verify the correctness. In cases where annotations are erroneous, the information is manually rectified by 7 well-informed analysts. The whole work took around 2.5 months to accomplish by 10 professional data annotators.

\subsection{Dataset Splits and Statistics}

In total, there are 1,446,500 face images from 25,000 subjects in the CAFR dataset. Each subject has 57.86 face images on average. The statistical comparisons between our CAFR and existing cross-age datasets are summarized in Tab.~\ref{tab: Table1}. CAFR is the largest and most comprehensive benchmark dataset for age-invariant face recognition to date. Following random selection, we divide the data into 10 splits with a pair-wise disjoint of subjects in each split. Each split contains 2,500 subjects and we randomly generate 5 genuine and 5 imposter pairs for each subject with various age gaps, resulting in 25,000 pairs per split. The remained data are preserved for algorithm development and parameter selection. We suggest evaluation systems to report the average \textbf{Acc}uracy (Acc), \textbf{E}qual \textbf{E}rror \textbf{R}ate (EER), \textbf{A}rea \textbf{U}nder the \textbf{C}urve (AUC) and \textbf{R}eceiver \textbf{O}perating \textbf{C}haracteristic (ROC) curve as 10-fold cross validation.

\section{Experiments}

We evaluate AIM qualitatively and quantitatively under
various settings for face recognition in the wild. In particular, we evaluate age-invariant face recognition performance on the CAFR dataset proposed in this work, as well as the MORPH~\cite{ricanek2006morph}, CACD~\cite{chen2015face} and FG-NET~\cite{fgnet} benchmark datasets. We also evaluate unconstrained face recognition results on the IJB-C benchmark dataset~\cite{maze2018iarpa} to verify the generalizability of AIM.

\paragraph{Implementation Details}

We apply \textbf{i}ntegrated \textbf{F}ace \textbf{A}nalytics \textbf{N}etwork (iFAN)~\cite{li2017integrated} for face \textbf{R}egion \textbf{o}f \textbf{I}nterest (RoI) extraction, 68 landmark localization (if not provided), and alignment; throughout the experiments, the sizes of the RGB image $x$, the attention score map $x^A$, the feature map $x^F$, the synthesized face image $\hat{x}$ are fixed as $128\times 128$; the pixel values of $x$, $\hat{x}$ and $x^R$ are normalized to [-1,1]; the sizes of the input local patches (w/o overlapping) to the discriminator $\mathrm{D}_{\phi_2}$ are fixed as $32\times 32$; the dimensionality of learned facial representation $f$ and sample $f^*$ drawn from prior distribution $U(f)$ are fixed as $256$; the age condition code $c$ is a 7-dimension one-hot vector to encode different age phases\footnote{We divide the whole age span into 7 age phases:  $\leq$20, 20-25, 25-30, 30-40, 40-50, 50-60, $\geq$60.}, based on which continuous face rejuvenation/aging results can be achieved through interpolation during inference; the element of $c$ is also confined to [-1,1], where -1 corresponds to 0; the element of smoothed labels for $\mathcal{L}_\mathrm{cer}$ is $\frac{1}{7}$; the constraint factors $\{\lambda_k\}_{k=_{1}^{14}}$ are empirically fixed as 0.1, 0.1, 0.01, 1.0, 0.01, 0.05, 0.1, $10^{-5}$, 0.03, 0.01, 0.05, 0.1, $10^{-5}$ and 0.03, respectively; the encoder $\mathrm{G}_{\theta_E}$ is initialized with the Light CNN-29~\cite{wu2018light} architecture by eliminating the linear classifier and replacing the activation function of the last fully-connected layer with hyperbolic tangent; the decoder $\mathrm{G}_{\theta_D}$ is initialized with 3 hidden fractionally-strided convolution layers with kernels $3\times 3\times 512 /2$, $3\times 3\times 256 /2$ and $3\times 3\times 128 /2$, activated with \textbf{Re}tified \textbf{L}inear \textbf{U}nit (ReLU)~\cite{dahl2013improving}, appended with a convolution layer with kernel $1\times 1\times 1$ activated with sigmoid and a convolution layer with kernel $1\times 1\times 3$ activated with scaled sigmoid for attention score map $x^A$ and feature map $x^F$ prediction, respectively; the domain classifier $\mathrm{C}_\varphi$ and the regularizer $\mathrm{R}_\psi$ are initialized with the same MLP architectures (which are learned separately), containing a hidden 256-way fully-connected layer activated with Leaky ReLU~\cite{maas2013rectifier} and a final 7-way fully-connected layer; the discriminator $\mathrm{D}_{\phi_1}$ is initialized with a MLP containing a hidden 256-way fully-connected layer activated with Leaky ReLU, appended with a 1-way fully-connected layer activated by sigmoid and a n-way fully-connected layer (n is the identity number of the training data) as the dual agents for $\mathcal{L}_\mathrm{{adv}_1}$ and $\mathcal{L}_\mathrm{ip}$, respectively; the discriminator $\mathrm{D}_{\phi_2}$ is initialized with a VGG-16~\cite{simonyan2014very} architecture by eliminating the linear classifier, and appending a new 1-way fully-connected layer activated by sigmoid and a new 7-way fully-connected layer activated by hyperbolic tangent as the dual agents for $\mathcal{L}_\mathrm{{adv}_2}$ and $\mathcal{L}_\mathrm{ae}$, respectively; the newly added layers are randomly initialized by drawing weights from a zero-mean Gaussian distribution with standard deviation 0.01; Batch Normalization~\cite{ioffe2015batch} is adopted in $\mathrm{G}_{\theta_E}$ and $\mathrm{G}_{\theta_D}$; the dropout~\cite{dahl2013improving} ratio is empirically fixed as 0.7; the weight decay and batch size are fixed as $5\times 10^{−3}$ and 32, respectively; We use an initial learning rate of $10^{-5}$ for pre-trained layers, and 2$\times$$10^{-4}$ for newly added layers in all our experiments; we decrease the learning rate to $\frac{1}{10}$ of the previous one after 20 epochs and train the network for roughly 60 epochs one after another; the proposed network is implemented based on the publicly available TensorFlow~\cite{abadi2016tensorflow} platform, which is trained using Adam ($\alpha$=2$\times$$10^{-4}$, $\beta_1$=0.5) on two NVIDIA GeForce GTX TITAN X GPUs with 12G memory; the same training setting is utilized for all our compared network variants.

\subsection{Evaluations on the CAFR Benchmark}

Our newly proposed CAFR dataset is the largest and most comprehensive age-invariant face recognition benchmark to date, which contains 1,446,500 images annotated with age, identity, gender, race and landmarks. Examples are visualized in Fig.~\ref{fig: Figure3}. The data are randomly organized into 10 splits, each consisting of 25,000 verification pairs with various age variations. Evaluation systems report Acc, EER, AUC and ROC as 10-fold cross validation.

\subsubsection{Component Analysis and Quantitative Comparison}

\begin{table}[t]
	\newcommand{\tabincell}[2]{\begin{tabular}{@{}#1@{}}#2\end{tabular}}
	\small
	\caption{\small Face recognition performance comparison on CAFR. The results are averaged over 10 testing splits.}
	\vspace{-5mm}
	\begin{center}
		\begin{tabular}{cccc}
			\toprule
			\tabincell{c}{\textbf{Model}} & \tabincell{c}{\textbf{Acc (\%)}} & \tabincell{c}{\textbf{EER (\%)}} & \tabincell{c}{\textbf{AUC (\%)}} \\
			\midrule
			Light CNN & 73.56$\pm$1.39 & 31.62$\pm$1.68 & 75.96$\pm$1.63 \\
			\cite{wu2018light}  & & & \\ 
			\midrule
			\multicolumn{4}{l}{\textbf{Architecture ablation of AIM }}     \\
			w/o $\mathrm{C}_\varphi$ & 78.85$\pm$1.39 & 21.97$\pm$1.18 & 86.77$\pm$1.01 \\
			w/o $\mathrm{R}_\psi$ & 80.39$\pm$1.19 & 20.22$\pm$1.25 & 88.52$\pm$0.82 \\
			w/o Att. & 82.25$\pm$1.03 & 18.50$\pm$1.04 & 90.26$\pm$0.94 \\
			\midrule
			\multicolumn{4}{l}{\textbf{Training loss ablation of AIM}}     \\
			w/o $\mathcal{L}_\mathrm{ip}$ & 67.64$\pm$0.88 & 45.85$\pm$2.59 & 57.14$\pm$2.59 \\
			w/o $\mathcal{L}_\mathrm{{adv}_1}$ & 81.02$\pm$1.10 & 19.56$\pm$1.00 & 89.10$\pm$0.83 \\
			w/o $\mathcal{L}_\mathrm{ae}$ & 81.83$\pm$1.29 & 19.08$\pm$1.03 & 89.87$\pm$0.76 \\
			w/o $\mathcal{L}_\mathrm{mc}$ & 82.03$\pm$0.98 & 18.57$\pm$0.98 & 90.10$\pm$0.83 \\
			w/o $\mathcal{L}_\mathrm{{adv}_2}$ & 82.30$\pm$0.99 & 18.28$\pm$1.02 & 90.32$\pm$0.71 \\
			\midrule
			\textbf{AIM} & \textbf{84.81$\pm$0.93} & \textbf{17.67$\pm$0.90} & \textbf{90.84$\pm$0.78} \\
			\bottomrule
		\end{tabular}
	\end{center}
	\label{tab: Table2}
\end{table}

\begin{figure*}[t]
	\begin{center}
		\includegraphics[width=1\linewidth]{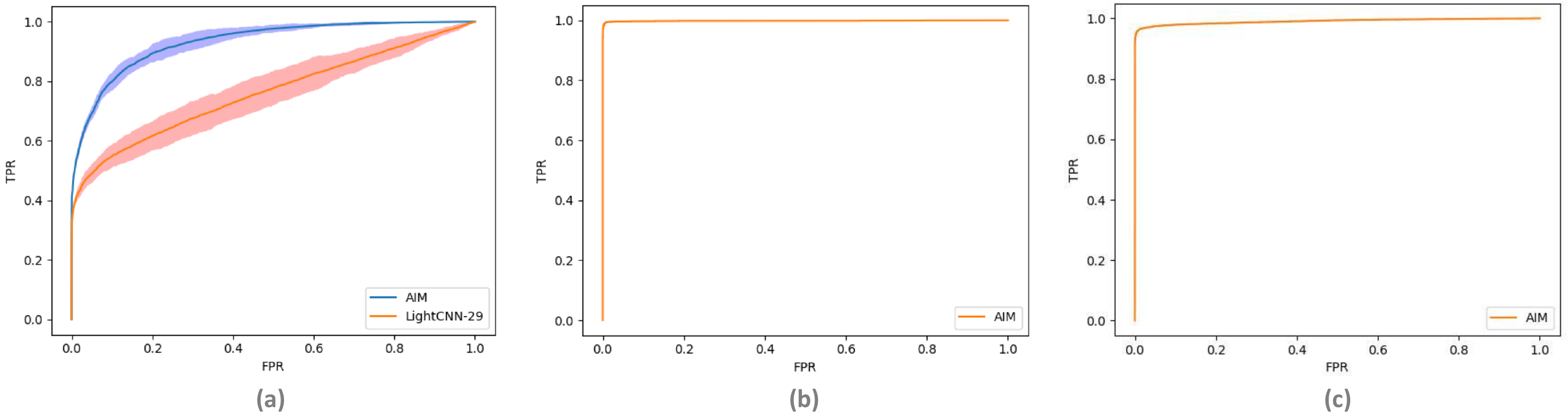}
	\end{center}
	\vspace{-6mm}
	\caption{\small ROC performance curve on (a) CAFR; (b) CACD-VS; (c) IJB-C. Best viewed in color.}
	\label{fig: FigureN}
\end{figure*}

We first investigate different architectures and loss function combinations of AIM to see their respective roles in age-invariant face recognition. We compare 10 variants from four aspects: baseline (Light CNN-29~\cite{wu2018light}), different network structures (w/o $\mathrm{C}_\varphi$, $\mathrm{R}_\psi$, w/o attention mechanism), different loss function combinations (w/o $\mathcal{L}_\mathrm{ip}$, $\mathcal{L}_\mathrm{{adv}_1}$, $\mathcal{L}_\mathrm{ae}$, $\mathcal{L}_\mathrm{mc}$, $\mathcal{L}_\mathrm{{adv}_2}$), and our proposed AIM.

The performance comparison w.r.t. Acc, EER and AUC on CAFR is reported in Tab.~\ref{tab: Table2}. The corresponding ROC curve is provided in Fig.~\ref{fig: FigureN} (a). By comparing the results from the $1^{st}$ \emph{v.s.} $4^{th}$ panels, we observe that our AIM consistently outperforms the baseline by a large margin: $11.25\%$ in Acc, $13.95\%$ in EER, and $14.88\%$ in AUC. Light-CNN is a general-purpose face recognition model, with representations entangled with age variations and suffering difficulties to distinguish cross-age faces. Comparatively, AIM jointly performs disentangled representation learning through cross-age domain adversarial training and cross-entropy regularization, and photorealistic cross-age face synthesis with attention mechanism in a mutual boosting way. By comparing the results from the $2^{nd}$ \emph{v.s.} $4^{th}$ panels, we observe that AIM consistently outperforms the 3 variants in terms of network structure. In particular, w/o $\mathrm{C}_\varphi$ refers to truncating the domain classifier from AIM, leading to $5.96\%$, $4.30\%$ and $4.07\%$ performance drop for all metrics. This verifies the necessity of cross-age domain adversarial training, which promotes encoded features to be indistinguishable w.r.t. the shift between multi-age domains to facilitate age-invariant representation learning. w/o $\mathrm{R}_\psi$ refers to truncating the cross-entropy regularizer from AIM, leading to $4.42\%$, $2.55\%$ and $2.32\%$ performance drop for all metrics. This verifies the necessity of cross-entropy regularization with label smoothing strategy that constrains cross-age representations with ambiguous separability to serve as an auxiliary assistance for $\mathrm{C}_\varphi$. The superiority of incorporating attention mechanism to cross-age face synthesis can be verified by comparing w/o Att. with AIM, \emph{i.e.}, $2.56\%$, $0.83\%$ and $0.58\%$ differences for all metrics. Identity-preserving quality is crucial for face recognition applications, the superiority of which is verified by comparing w/o $\mathcal{L}_\mathrm{ip}$ with AIM, \emph{i.e.}, $17.17\%$, $28.18\%$ and $33.70\%$ decline for all metrics. The superiority of incorporating adversarial learning to specific process can be verified by comparing w/o $\mathcal{L}_\mathrm{{adv}_i}$, $i\in \{1,2\}$ with AIM, \emph{i.e.}, $3.79\%$, $1.89\%$ and $1.74\%$; $2.51\%$, $0.61\%$ and $0.52\%$ decrease for all metrics. The superiorities of incorporating age estimation and manifold consistency constraints are verified by comparing w/o $\mathcal{L}_\mathrm{ae}$ and w/o $\mathcal{L}_\mathrm{mc}$ with AIM, \emph{i.e.}, $2.98\%$, $1.41\%$ and $0.97\%$; $2.78\%$, $0.90\%$ and $0.74\%$ drop for all metrics.

\subsubsection{Qualitative Comparison}

\begin{figure}[t]
	\begin{center}
		\includegraphics[width=1\linewidth]{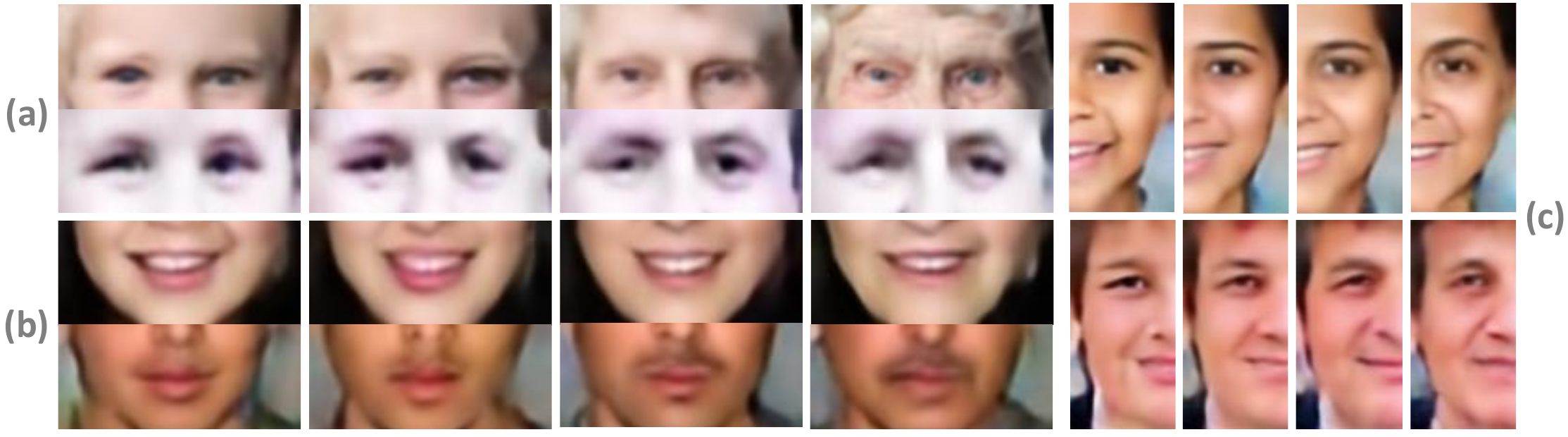}
	\end{center}
	\vspace{-6mm}
	\caption{\small Facial attributes transformation over time in terms of (a) wrinkles \& eyes, (b) mouth \& moustache and (c) laugh lines, which is automatically learned by AIM instead of physical modelling.}
	\label{fig: Figure6}
\end{figure}

\begin{figure}[t]
	\begin{center}
		\includegraphics[width=1\linewidth]{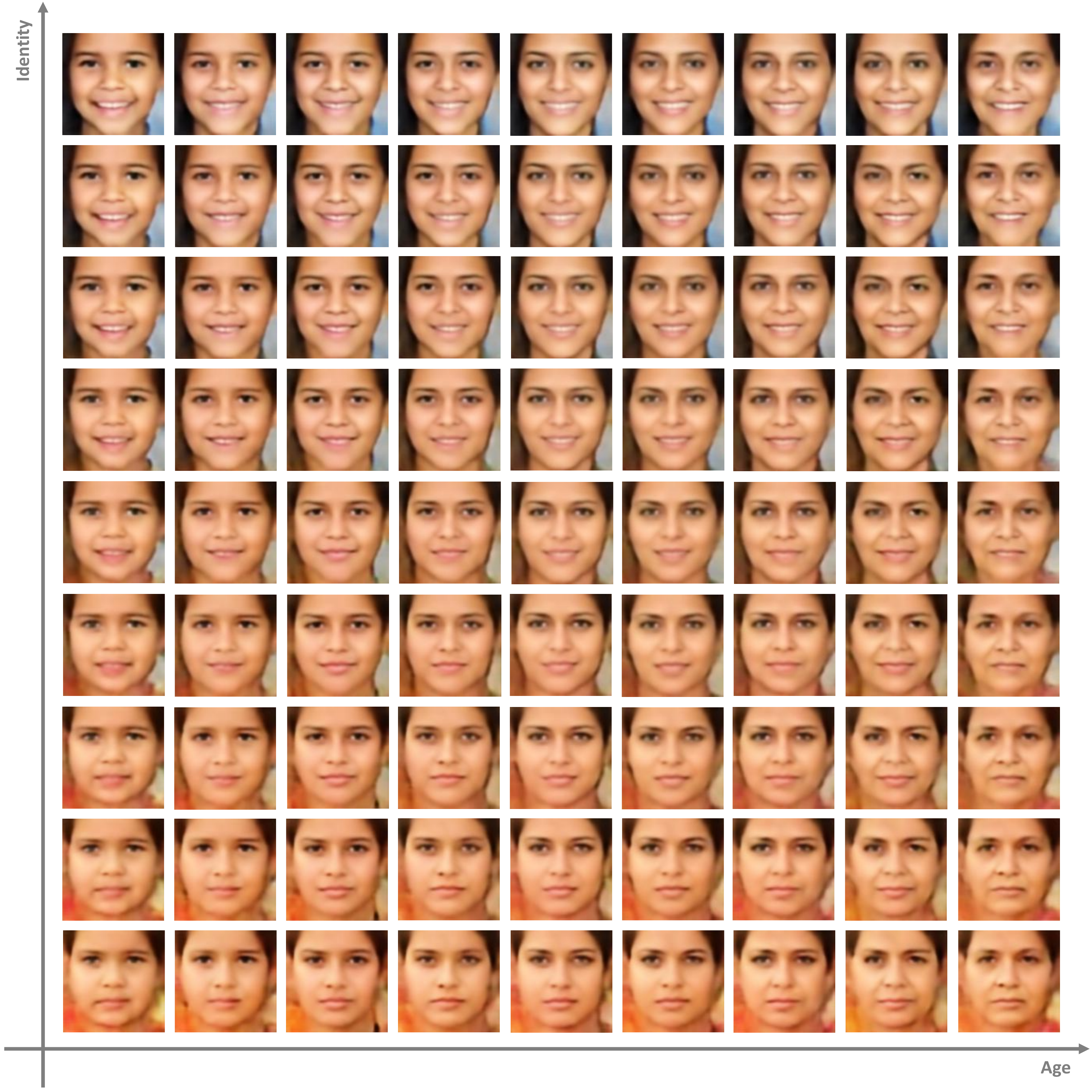}
	\end{center}
	\vspace{-6mm}
	\caption{\small Illustration of learned face manifold with continuous transitions  in age (horizontal axis) and identity (vertical axis).}
	\label{fig: Figure7}
\end{figure}

\begin{figure*}[t]
	\begin{center}
		\includegraphics[width=1\linewidth]{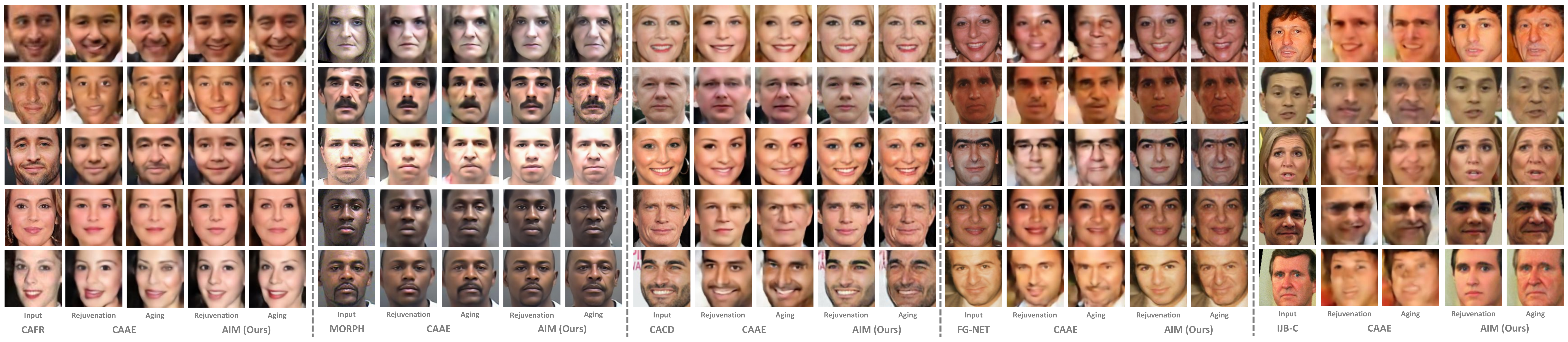}
	\end{center}
	\vspace{-6mm}
	\caption{\small Qualitative comparison of face rejuvenation/aging results on CAFR, MORPH, CACD, FG-NET and IJB-C.}
	\label{fig: Figure_en}
\end{figure*}

\begin{figure*}[t]
	\begin{center}
		\includegraphics[width=1\linewidth]{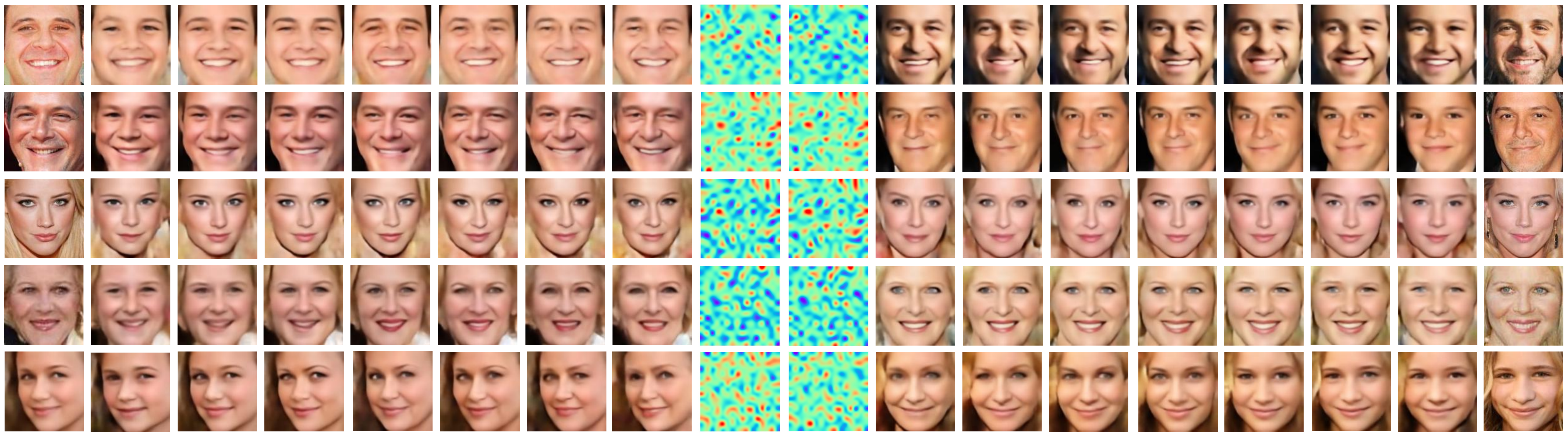}
	\end{center}
	\vspace{-6mm}
	\caption{\small Age-invariant face recognition example results on CAFR. \emph{Col.} 1 \& 18: Input faces of distinct identities with various challenging factors (\emph{e.g.}, neutral, illumination, expression, and pose). \emph{Col.} 2, 3, 4, 5, 6, 7, 8, 11, 12, 13, 14, 15, 16, 17: Synthesized age regressed/progressed faces by our proposed AIM. \emph{Col.} 9 \& 10: Learned facial representations by  AIM, which are explicitly disentangled from age variations. These examples indicate facial representations learned by AIM are robust to age variance, and synthesized cross-age face images retain the intrinsic details. Best viewed in color.}
	\label{fig: Figure4}
\end{figure*}

Most previous works on age-invariant face recognition address this problem considering either only robust representation learning or only face rejuvenation/aging. It is commonly believed simultaneously modeling both is a highly non-linear transformation, thus it is difficult for a model to learn discriminative and age-invariant facial representations while generating faithful cross-age face images. However, with enough training data and proper architecture and objective function design of  AIM, it is feasible to take the best of both worlds, as shown in Fig.~\ref{fig: Figure1}. For more detailed results across a wide range of ages in high resolution, please refer to Fig.~\ref{fig: Figure4}. Our AIM consistently provides discriminative and age-invariant representations and high-fidelity age regressed/progressed face images for all cases. This well verifies that the joint learning scheme of age-invariant representation and attention-based cross-age face synthesis is effective, and both results are beneficial to face recognition in the wild.

We then visually compare the qualitative face rejuvenation and aging results by our AIM with previous state-of-the-art method CAAE~\cite{zhifei2017cvpr} in Fig.~\ref{fig: Figure_en} $1^{st}$ block and showcase the facial detail transformation over time with AIM in Fig.~\ref{fig: Figure6}. It can be observed that AIM achieves simultaneous face rejuvenation and aging with photorealistic and accurate age transformation effect (\emph{e.g.}, wrinkles, eyes, mouth, moustache, laugh lines), thanks to the novel network structure and training strategy. In contrast, results of previous work may suffer from blur and ghosting artifacts, and be fragile to variations in illumination, expression and pose. This further shows effectiveness of the proposed AIM.

To demonstrate the capacity of AIM to synthesize cross-age face images with continuous and smooth transition between identities and ages, and show that the learned representations are identity-specific and explicitly disentangled from age variations, we further visualize the learned face manifold in Fig.~\ref{fig: Figure7} by performing interpolation upon both $f$ and $c$. In particular, we take two images of different subjects $x_1$ and $x_2$, extract the encoded features from $\mathrm{G}_{\theta_E}$ and perform interpolation between $f_{x_1}$ and $f_{x_2}$. We also interpolate between two neighboring age condition codes to generate face images with continuous ages. The interpolated $f$ and $c$ are then fed to $\mathrm{G}_{\theta_D}$ to synthesize face images. These smooth semantic changes indicate that the model has learned to produce identity-specific representations disentangled from age variations for age-invariant face recognition.

\begin{figure}[t]
	\begin{center}
		\includegraphics[width=0.8\linewidth]{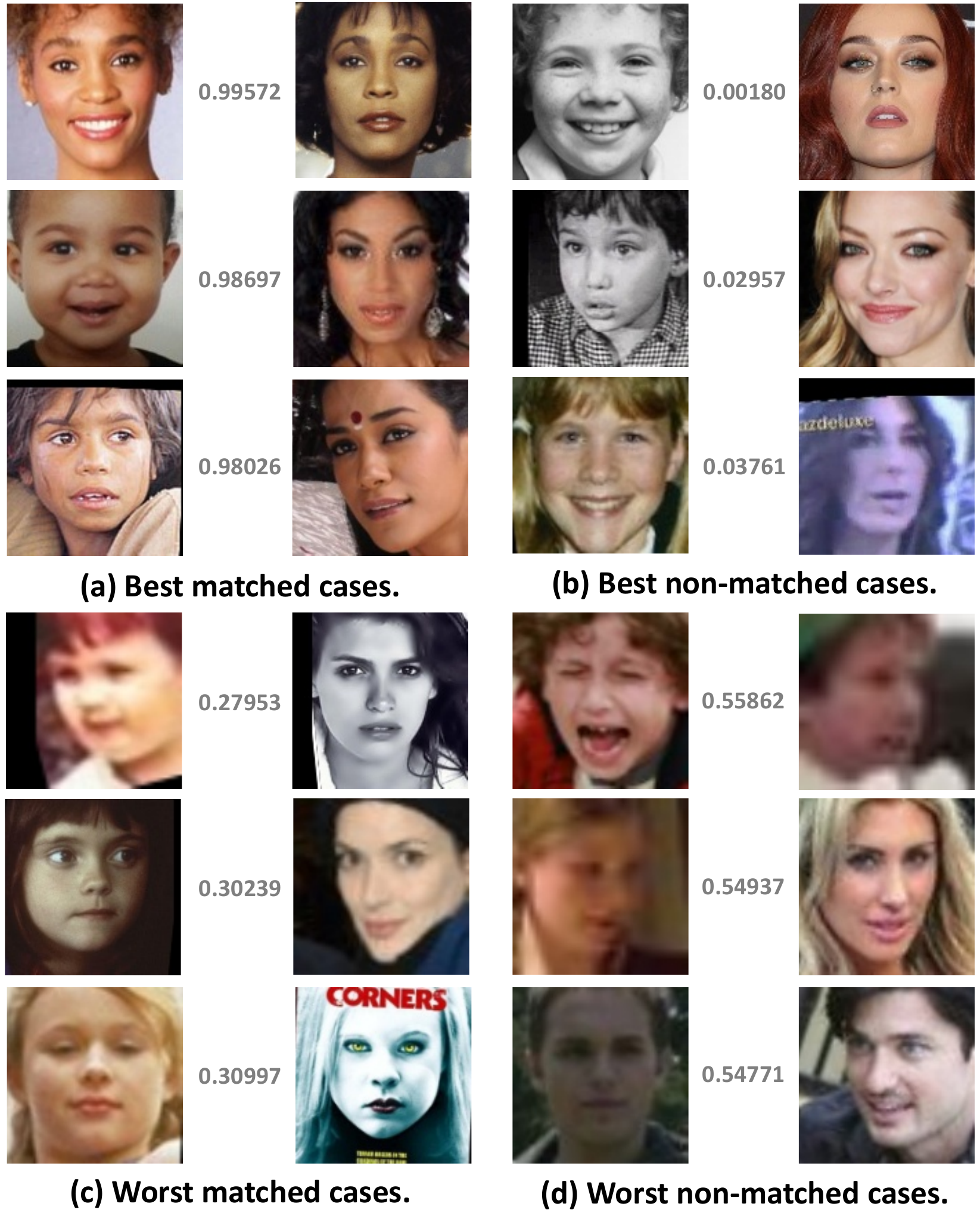}
	\end{center}
	\vspace{-6mm}
	\caption{\small Age-invariant face recognition analysis on CAFR split1.}
	\label{fig: Figure8}
\end{figure}

Finally, we visualize the cross-age face verification results for CAFR split1 to gain insights into age-invariant face recognition with AIM. After computing the similarities for all pairs of probe and reference sets, we sort the results into a ranking list. Each row shows a probe and reference pair. Between pairs are the matching similarities. Fig.~\ref{fig: Figure8} (a) and (b) show the best matched and non-matches examples, respectively. We note that most of these cases are under mild conditions in terms of age gap and other unconstrained factors like resolution, expression and pose. Fig.~\ref{fig: Figure8} (c) and (d) show the worst matched and non-matched examples, respectively, representing failed matching. We note that most of error cases are with large age gaps blended with other challenging scenarios like blur, extreme expressions, heavy make-up and large poses, which are even hard for humans to recognize.  This confirms that CAFR aligns well with reality and deserves more research attention.

\subsection{Evaluations on the MORPH Benchmark}

\begin{table}[t]
	\newcommand{\tabincell}[2]{\begin{tabular}{@{}#1@{}}#2\end{tabular}}
	\begin{center}
		\small
		\caption{\small Rank-1 recognition rates (\%) on MORPH Album2.}
		\label{tab: Table3}
		\vspace{-2mm}
		\begin{tabular}{cc}
			\hline
			{\textbf{Method}} & {\textbf{Setting-1/Setting-2}} \\
			\hline
			HFA~\cite{gong2013hidden} & 91.14/- \\
			CARC~\cite{chen2014cross} & 92.80/- \\
			MEFA~\cite{gong2015maximum} & 93.80/- \\
			GSM~\cite{lin2017cross} & -/94.40 \\
			MEFA+SIFT+MLBP~\cite{gong2015maximum} & 94.59/- \\
			LPS+HFA~\cite{li2016aging} & 94.87/- \\
			LF-CNN~\cite{wen2016latent} & 97.51/- \\
			AE-CNN~\cite{zheng2017age} & -/98.13 \\
			OE-CNN~\cite{wangorthogonal} & 98.55/98.67 \\
			\hline
			\textbf{AIM (Ours)} & \textbf{99.13}/\textbf{98.81} \\
			\textbf{AIM + CAFR (Ours)} & \textbf{99.65}/\textbf{99.26} \\
			\hline
		\end{tabular}
	\end{center}
\end{table}

MORPH is a large-scale public longitudinal face database, collected in real-world conditions with variations in age, pose, expression and lighting conditions. It has two separate datasets: Album1 and Album2. Album 1 contains 1,690 face images from 515 subjects while Album 2 contains 78{,}207 face images from 20{,}569 subjects. Statistical details are provided in Tab.~\ref{tab: Table1}. Both albums include meta data for age, identity, gender, race, eye coordinates and date of acquisition. For fair comparisons, Album2 is used for evaluation. Following~\cite{li2011discriminative,gong2013hidden}, Album2 is partitioned into a training set of 20{,}000 face images from 10{,}000 subjects with each subject represented by two images with largest gap, and an independent testing set consisting of a gallery set and a probe set from the remaining subjects under two settings. Setting-1 consists of 20{,}000 face images from 10{,}000 subjects with each subject represented by a youngest face image as gallery and an oldest face image as probe while Setting-2 consists of 6{,}000 face images from 3{,}000 subjects with the same criteria. Evaluation systems report the Rank-1 identification rate.

The face recognition performance comparison of the proposed AIM with other state-of-the-arts on MORPH~\cite{ricanek2006morph} Album2 in Setting-1 and Setting-2 is reported in Tab.~\ref{tab: Table3}. With the mutual boosting learning scheme of age-invariant representation and attention-based cross-age face synthesis, our method outperforms the $2^{nd}$-best by 0.58\% and 0.14\% for Setting-1 and Setting-2, respectively. By incorporating CAFR during training, the rank-1 recognition rates are further improved by 0.52\% and 0.45\% for Setting-1 and Setting-2, respectively. This confirms that our AIM is highly effective and the proposed CAFR dataset is beneficial for advancing age-invariant face recognition performance. Visual comparison of face rejuvenation/aging results by AIM and CAAE~\cite{zhifei2017cvpr} is provided in Fig.~\ref{fig: Figure_en} $2^{nd}$ block, also validating advantages of AIM over existing solutions.

\subsection{Evaluations on the CACD Benchmark}

\begin{table}[t]
	\newcommand{\tabincell}[2]{\begin{tabular}{@{}#1@{}}#2\end{tabular}}
	\begin{center}
		\small
		\caption{\small Face recognition performance comparison on CACD-VS.}
		\label{tab: Table4}
		\vspace{-3mm}
		\begin{tabular}{cc}
			\hline
			{\textbf{Method}} & {\textbf{Acc (\%)}} \\
			\hline
			CAN~\cite{xu2017age} & 92.30 \\
			VGGFace~\cite{parkhi2015deep} & 96.00 \\
			Center Loss~\cite{wen2016discriminative} & 97.48 \\
			MFM-CNN~\cite{wu2018light} & 97.95 \\
			LF-CNN~\cite{wen2016latent} & 98.50 \\
			Marginal Loss~\cite{deng2017marginal} & 98.95 \\
			DeepVisage~\cite{hasnat2017deepvisage} & 99.13 \\
			OE-CNN~\cite{wangorthogonal} & 99.20 \\
			\hline
			Human, \emph{avg.}~\cite{chen2015face} & 85.70 \\
			Human, voting~\cite{chen2015face} & 94.20 \\
			\hline
			\textbf{AIM (Ours)} & \textbf{99.38} \\
			\textbf{AIM + CAFR (Ours)} & \textbf{99.76} \\
			\hline
		\end{tabular}
	\end{center}
\end{table}

CACD is a large-scale public dataset for face recognition and retrieval across ages, with variations in age, illumination, makeup, expression and pose, aligned with the real-world scenarios better than MORPH~\cite{ricanek2006morph}. It contains 163,446 face images from 2,000 celebrities. Statistical details are provided in Tab.~\ref{tab: Table1}. The meta data include age, identity and landmark. However, CACD contains some incorrectly labeled samples and  duplicate images. For fair comparison, following~\cite{chen2015face}, a carefully annotated version \textbf{CACD} \textbf{V}erification \textbf{S}ub-set (CACD-VS) is used for evaluation. It consists of 10 splits including 4,000 image pairs in total. Each split contains 200 genuine pairs and 200 imposter pairs for cross-age verification task. Evaluation systems report Acc and ROC as 10-fold cross validation.

The face recognition performance comparison of the proposed AIM with other state-of-the-arts on CACD-VS~\cite{chen2015face} is reported in Tab.~\ref{tab: Table4}. The corresponding ROC curve is provided in Fig.~\ref{fig: FigureN} (b). Our method dramatically surpasses human performance and other state-of-the-arts. In particular, AIM improves the Acc of the $2^{nd}$-best by 0.18\%. AIM also outperforms human voting performance by 5.18\%. To our best knowledge, this is the new state-of-the-art, including unpublished technical reports. This shows the learned facial representations by AIM are discriminative and robust even with in-the-wild variations. With the injection of CAFR as augmented training data, our method further gains 0.38\%. Visual comparison of face rejuvenation/aging results by AIM and four state-of-the-art methods is provided in Fig.~\ref{fig: Figure_en} $3^{rd}$ block, which again verifies effectiveness of our method for high-fidelity cross-age face synthesis.

\subsection{Evaluations on the FG-NET Benchmark}

\begin{table}[t]
	\newcommand{\tabincell}[2]{\begin{tabular}{@{}#1@{}}#2\end{tabular}}
	\small
	\caption{\small Face recognition performance comparison on FG-NET.}
	\vspace{-5mm}
	\begin{center}
		\begin{tabular}{ccc}
			\hline
			\tabincell{c}{\textbf{Method}} & \tabincell{c}{\textbf{Rank-1 (\%)}} \\
			\hline
			Park \emph{et al.}~\cite{park2010age} & 37.40 \\
			Li \emph{et al.}~\cite{li2011discriminative} & 47.50 \\
			HFA~\cite{gong2013hidden} & 69.00 \\
			MEFA~\cite{gong2015maximum} & 76.20 \\
			CAN~\cite{xu2017age} & 86.50 \\
			LF-CNN~\cite{wen2016latent} & 88.10 \\
			\hline
			\textbf{AIM (Ours)} & \textbf{93.20} \\
			\hline
		\end{tabular}
	\end{center}
	\label{tab: Table5}
\end{table}

FG-NET is a popular public dataset for cross-age face recognition, collected in realistic conditions with huge variability in age covering from child to elder. It contains 1{,}002 face images from 82 non-celebrity subjects. Statistical details are provided in Tab.~\ref{tab: Table1}. The meta data include age, identity and landmark. Since the size of FG-NET is small, we follow the leave-one-out setting of~\cite{li2011discriminative,gong2013hidden} for fair comparisons with previous methods. In particular, we leave one image as the testing sample and train (finetune) the model with remaining 1{,}001 images. We repeat this procedure 1{,}002 times and report the average rank-1 recognition rate.

The face recognition performance comparison of the proposed AIM with other state-of-the-arts on FG-NET~\cite{fgnet} is reported in Tab.~\ref{tab: Table5}. AIM improves the $2^{nd}$-best by 5.10\%. Qualitative comparisons for face rejuvenation/aging are provided in Fig.~\ref{fig: Figure_en} $4^{th}$ block, which well shows the promising potential of our method for challenging unconstrained face recognition contaminated with age variance.

\subsection{Evaluations on the IJB-C Benchmark}

\begin{table*}[t]
	\newcommand{\tabincell}[2]{\begin{tabular}{@{}#1@{}}#2\end{tabular}}
	\begin{center}
		\small
		\caption{\small Face recognition performance comparison on IJB-C.}
		\label{tab: Table6}
		\vspace{-2mm}
		\begin{tabular}{ccccc}
			\hline
			{\textbf{Method}} & \tabincell{c}{\textbf{TAR@FAR=$10^{-5}$}} & \tabincell{c}{\textbf{TAR@FAR=$10^{-4}$}} & \tabincell{c}{\textbf{TAR@FAR=$10^{-3}$}} & \tabincell{c}{\textbf{TAR@FAR=$10^{-2}$}} \\
			\hline
			GOTS~\cite{maze2018iarpa} & 0.066 & 0.147 & 0.330 & 0.620 \\
			FaceNet~\cite{Schroff:Facenet} & 0.330 & 0.487 & 0.665 & 0.817 \\
			VGGFace~\cite{parkhi2015deep} & 0.437 & 0.598 & 0.748 & 0.871 \\
			VGGFace2\_ft~\cite{cao2018vggface2} & 0.768 & 0.862 & 0.927 & 0.967 \\
			MN-vc~\cite{xie2018multicolumn} & 0.771 & 0.862 & 0.927 & 0.968 \\
			\hline
			\textbf{AIM} & \textbf{0.826} & \textbf{0.895} & \textbf{0.935} & \textbf{0.962} \\
			\hline
		\end{tabular}
	\end{center}
\end{table*}

IJB-C contains 31{,}334 images and 11{,}779 videos from 3{,}531 subjects, which are split into 117{,}542 frames, 8.87 images and 3.34 videos per subject, captured from in-the-wild environments to avoid the near frontal bias. For fair comparison, we follow the template-based setting and evaluate models on the standard 1:1 verification protocol in terms of \textbf{T}rue \textbf{A}cceptance \textbf{R}ate (TAR)@\textbf{F}alse \textbf{A}cceptance \textbf{R}ate (FAR).

The face recognition performance comparison of the proposed AIM with other state-of-the-arts on IJB-C~\cite{maze2018iarpa} unconstrained face verification protocol is reported in Tab.~\ref{tab: Table6}. The corresponding ROC curve is provided in Fig.~\ref{fig: FigureN} (c). Our AIM beats the $2^{nd}$-best by 5.50\% in TAR@FAR=$10^{-5}$, which verifies its remarkable generalizability for recognizing faces in the wild. Qualitative comparisons for face rejuvenation/aging are provided in Fig.~\ref{fig: Figure_en} $5^{th}$ block, which further shows the superiority of our method for cross-age face synthesis under unconstrained condition.

\section{Conclusion}

We proposed a novel \textbf{A}ge-\textbf{I}nvariant \textbf{M}odel (AIM) for joint disentangled representation learning and photorealistic cross-age face synthesis to address the challenging face recognition with large age variations. Through carefully designed network architecture and optimization strategies, AIM learns to generate powerful age-invariant facial representations explicitly disentangled from the age variation while achieving continuous face rejuvenation/aging with remarkable photorealistic and identity-preserving properties, avoiding requirements of paired data and true age of testing samples. Moreover, we propose a new large-scale \textbf{C}ross-\textbf{A}ge \textbf{F}ace \textbf{R}ecognition (CAFR) dataset to spark progress in age-invariant face recognition. Comprehensive experiments demonstrate the superiority of AIM over the state-of-the-arts. We envision the proposed method and benchmark dataset would drive the age-invariant face recognition research towards real-world applications with presence of age gaps and other complex unconstrained distractors.

\section*{Acknowledgement}

The work of Jian Zhao was partially supported by \textbf{C}hina \textbf{S}cholarship \textbf{C}ouncil (CSC) grant 201503170248. 

The work of Junliang Xing was partially supported by the National Science Foundation of Chian 61672519.

The work of Jiashi Feng was partially supported by NUS IDS R-263-000-C67-646, ECRA R-263-000-C87-133 and MOE Tier-II R-263-000-D17-112.

{\small
\bibliographystyle{ieee}
\bibliography{egbib}
}

\end{document}